\documentclass[runningheads]{llncs}
\usepackage{graphicx}

\usepackage{tikz}
\usepackage{comment}
\usepackage{amsmath,amssymb} 
\usepackage{color}

\usepackage[accsupp]{axessibility}  

\usepackage{multirow}
\usepackage{graphicx}
\usepackage{hyperref}

\def\ie{{\textit{i.e.}}}
\def\eg{{\textit{e.g.}}}
\def\etc{{\textit{etc}}}

\makeatletter
\def\thanks#1{\protected@xdef\@thanks{\@thanks
        \protect\footnotetext{#1}}}
\makeatother
\usepackage[marginal]{footmisc}
\usepackage[misc]{ifsym}

\usepackage{amsmath,amsfonts,bm}

\def\eqref#1{equation~\ref{#1}}

\def\1{\bm{1}}

\newcommand{\test}{\mathcal{D_{\mathrm{test}}}}

\def\vp{{\bm{p}}}

\def\vu{{\bm{u}}}

\def\vx{{\bm{x}}}

\def\mA{{\bm{A}}}

\def\mD{{\bm{D}}}

\def\mL{{\bm{L}}}

\def\mP{{\bm{P}}}

\def\mU{{\bm{U}}}

\def\mLambda{{\bm{\Lambda}}}

\DeclareMathAlphabet{\mathsfit}{\encodingdefault}{\sfdefault}{m}{sl}
\SetMathAlphabet{\mathsfit}{bold}{\encodingdefault}{\sfdefault}{bx}{n}

\def\gE{{\mathcal{E}}}

\def\gG{{\mathcal{G}}}

\def\gV{{\mathcal{V}}}

\begin{document}
\pagestyle{headings}
\mainmatter
\def\ECCVSubNumber{4378}  

\title{Exploring the Devil in Graph Spectral Domain for 3D Point Cloud Attacks} 


\titlerunning{Exploring the Devil in Graph Spectral Domain for 3D Point Cloud Attacks}
%
\author{Qianjiang Hu\textsuperscript{*} \and
Daizong Liu\textsuperscript{*}\and
Wei Hu\textsuperscript{\Letter}
\thanks{
\textsuperscript{*} Q. Hu and D. Liu contributed equally to this work.
\textsuperscript{\Letter} Corresponding author: W. Hu.
\\This work was supported by National Natural Science Foundation of China under Contract No. 61972009. }
}

\authorrunning{Q Hu, D Liu, et al.}
%
\institute{Wangxuan Institute of Computer Technology, Peking University\\
No. 128, Zhongguancun North Street, Beijing, China\\
\email{hqjpku@pku.edu.cn, dzliu@stu.pku.edu.cn, forhuwei@pku.edu.cn}}


\maketitle

\begin{abstract}
With the maturity of depth sensors, point clouds have received increasing attention in various applications such as autonomous driving, robotics, surveillance, \etc., while deep point cloud learning models have shown to be vulnerable to adversarial attacks.
Existing attack methods generally add/delete points or perform point-wise perturbation over point clouds to generate adversarial examples in the data space, which may neglect the geometric characteristics of point clouds.
Instead, we propose point cloud attacks from a new perspective---Graph Spectral Domain Attack (GSDA), aiming to perturb transform coefficients in the graph spectral domain that corresponds to varying certain geometric structure. 
In particular, we naturally represent a point cloud over a graph, and adaptively transform the coordinates of points into the graph spectral domain via graph Fourier transform (GFT) for compact representation. 
We then analyze the influence of different spectral bands on the geometric structure of the point cloud, based on which we propose to perturb the GFT coefficients in a learnable manner guided by an energy constraint loss function. 
Finally, the adversarial point cloud is generated by transforming the perturbed spectral representation back to the data domain via the inverse GFT (IGFT).
Experimental results demonstrate the effectiveness of the proposed GSDA in terms of both imperceptibility and attack success rates under a variety of defense strategies.
The code is available at \href{https://github.com/WoodwindHu/GSDA}{https://github.com/WoodwindHu/GSDA}.
\end{abstract}

\section{Introduction}
\label{sec:intro}
Deep Neural Networks (DNNs) are known to be vulnerable to adversarial examples \cite{szegedy2013intriguing,goodfellow2014explaining}, which are indistinguishable from legitimate ones by adding trivial perturbations, but lead to incorrect model prediction. Many efforts have been made into the attacks on the 2D image field \cite{dong2018boosting,madry2017towards,kurakin2016adversarial,tu2019autozoom}, which often add imperceptible pixel-wise noise onto images to deceive the DNNs. Nevertheless, adversarial attacks on 3D point clouds---discrete representations of 3D scenes or objects that consist of a set of points residing on irregular domains---are still relatively under-explored, which are however crucial in various safety-critical applications such as autonomous driving \cite{chen2017multi} and medical data analysis \cite{singh20203d}.

\begin{figure}[t]
\begin{center}
    \includegraphics[width=0.8\columnwidth]{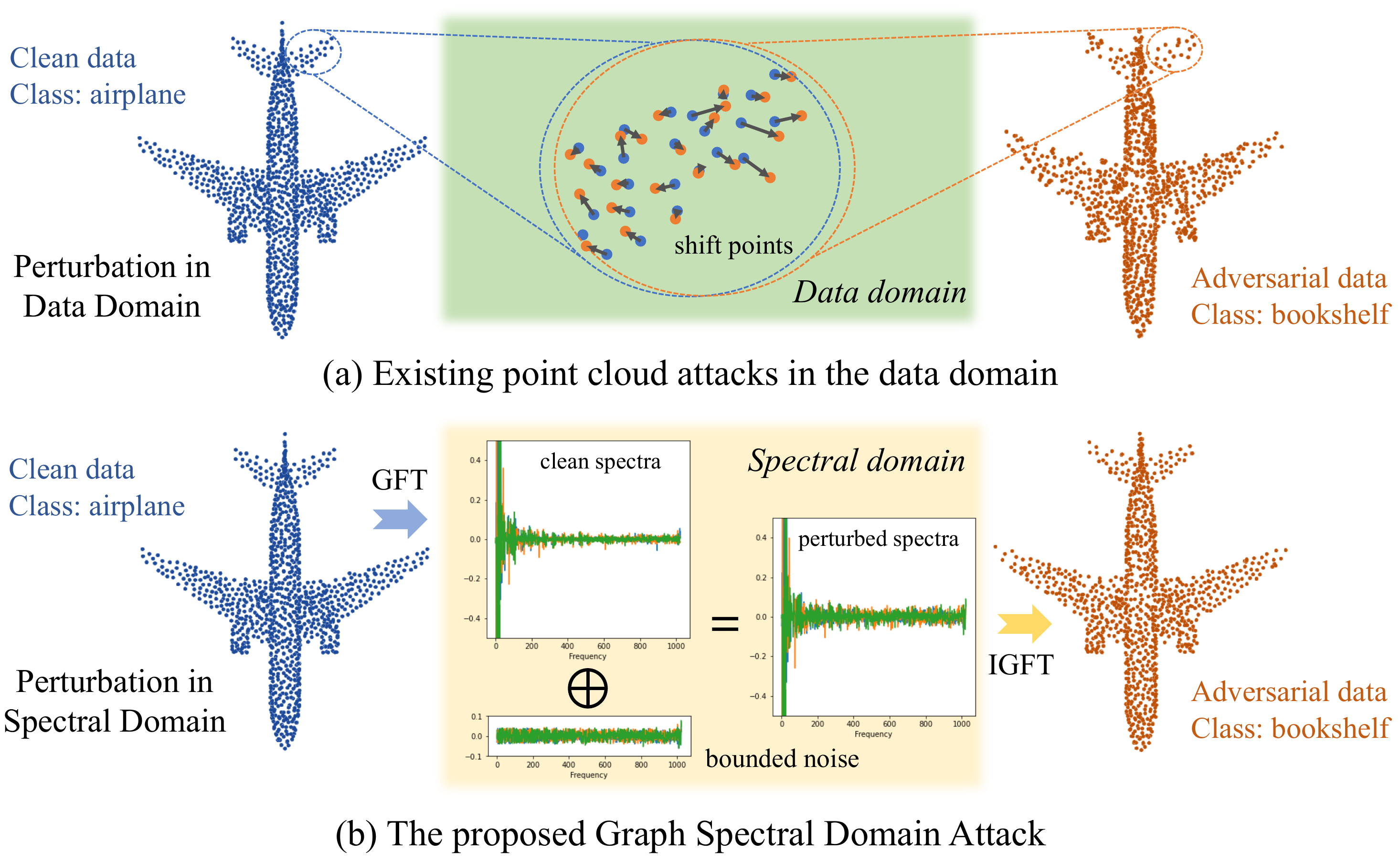}
\end{center}
\vspace{-14pt}
\caption{(a) Existing point cloud attacks generally perturb point coordinates by shifting points in the data domain. (b) We explore perturbation in the graph spectral domain, leading to more imperceptible and effective adversarial examples. 
This is enlightened by spectral characteristics: most energies are concentrated in low-frequency components that represent the rough shape of point clouds while high-frequency components encode fine-grained details.   
When attacked with bounded spectral noise, spectra of the point cloud keeps similar patterns with the clean one, thus preserving geometric structures.
}
\label{fig:teaser}
\vspace{-12pt}
\end{figure}
Existing 3D point cloud attacks \cite{xiang2019generating,wicker2019robustness,zhang2019adversarial,zheng2019pointcloud,tsai2020robust,zhao2020isometry,zhou2020lg,hamdi2020advpc} are all developed in the {\it data space}. Some of them \cite{xiang2019generating,zhang2019adversarial,wicker2019robustness,zheng2019pointcloud} employ the gradient search method to identify critical points from point clouds and modify (add or delete) them to distort the most representative features for misclassification. 
Recently, more works \cite{wen2020geometry,tsai2020robust,carlini2017towards,hamdi2020advpc,liu2019extending,ma2020efficient,zhang2019defense} follow the C\&W framework \cite{goodfellow2014explaining} to learn to perturb xyz coordinates of each point by gradient optimization in an end-to-end manner. 
Although the above two types of works achieve high attack success rates, the perturbed point clouds are often easily perceivable to humans, such as outliers and uneven distribution. 
This is because preserving geometric characteristics of point clouds is generally not considered yet in these methods.

To this end, we propose point cloud attacks from a new perspective---Graph Spectral Domain Attack (GSDA), aiming to exploit the elegant characterization of geometric structures in the graph spectral domain and thereby perturb graph transform coefficients that explicitly varies certain geometric structure. 
On the one hand, we provide graph spectral analysis of point clouds, which shows that the rough shape of point clouds is represented by low-frequency components while the fine details of objects are encoded in high-frequency components in general.
With trivial perturbation in the spectral domain, the point cloud could retain the original rough shape with similar local details, as shown in Figure~\ref{fig:teaser}.
On the other hand, the spectral characteristics of point clouds represent higher-level and global information than point-to-point relations leveraged in previous works. 
That is, the spectral representation encodes more abstract and essential contexts for recognizing the point cloud. 

Based on the above analysis, we develop a novel paradigm of Graph Spectral Domain Attack (GSDA) for point cloud attacks.
In particular, different from images that are sampled on regular grids and typically transformed in the Discrete Cosine Transform (DCT) domain, point clouds reside on irregular domains with no ordering of points. 
Hence, we represent a point cloud over a graph naturally and adaptively, where each point is treated as a vertex and connected to its $K$ nearest neighbors, and the coordinates of each point serve as the graph signal. 
Then, we compute the graph Laplacian matrix \cite{shuman2013emerging} that encodes the edge connectivity and vertex degree of the graph, whose eigenvectors form the basis of the Graph Fourier Transform (GFT) \cite{hammond2011wavelets}. 
Because of the compact representation of point clouds in the GFT domain \cite{hu2021overview}, we transform the coordinate signal of point clouds onto the spectral domain via the GFT, leading to transform coefficients corresponding to each spectral component. 
Next, we develop a learnable spectrum-aware perturbation approach to perturb the spectral domain with adversarial noise. 
In order to keep the energy balance among the whole frequency bands, we design an energy constraint function to restrict the perturbation size.
Finally, we transform the perturbed GFT coefficients back to the data domain via the inverse GFT (IGFT) to produce the crafted point cloud.
We iteratively optimize the adversarial loss function, and perform back-propagation to retrieve the gradient in the spectral domain for generating and updating the desired spectrum-aware perturbations. 
The point clouds reconstructed by the IGFT are taken as the adversarial examples.


Our main contributions are summarized as follows:
\vspace{-2pt}
\begin{itemize}
    \item We propose a novel paradigm of point cloud attacks---Graph Spectral Domain Attack (GSDA), which perturbs point clouds in the graph spectral domain to exploit the high-level spectral characterization of geometric structures. Such spectral approach marks the first significant departure from the current practice of point cloud attacks.    
    \item We provide in-depth graph spectral analysis of point clouds, which enlightens our formulation of exploring destructive perturbation on appropriate frequency components. 
    Based on this, we develop a learnable spectrum-aware perturbation approach to attack 3D models in an end-to-end manner.
    \item Extensive experiments show that the proposed GSDA achieves 100\% of attack success rates in both targeted and untargeted settings with the least required perturbation size. We also demonstrate the imperceptibility of the GSDA compared to state-of-the-arts, as well as the robustness of the GSDA by attacking existing defense methods and implementing transfer-based attacks. 
\end{itemize}

\section{Related Works}
\label{sec:relat}
\noindent \textbf{3D Point Cloud Classification.}
Deep 3D point cloud learning has emerged in recent years, which has diverse applications in many fields, such as 3D object classification \cite{su2015multi,yu2018multi}, 3D scene segmentation \cite{graham20183d,wang2018sgpn,xu2020grid}, and 3D object detection in autonomous driving \cite{chen2017multi,yang2019learning}. 
Among them, 3D object classification is the most fundamental yet important task, which learns representative information including both local details and global context of point clouds.
Early works attempt to classify point clouds by adapting deep learning models in the 2D space \cite{su2015multi,yu2018multi}.
In order to directly learn the 3D structure and address the unorderness problem of point clouds, 
pioneering methods DeepSets \cite{zaheer2017deep} and PointNet \cite{qi2017pointnet} propose to achieve end-to-end learning on point cloud classification by formulating a general specification.
PointNet++ \cite{qi2017pointnet++} and other extensive works \cite{duan2019structural,liu2019densepoint,yang2019modeling} are built upon PointNet to further capture the fine local structural information from the neighborhood of each point.
Recently, some works focus on either designing special convolutions on the 3D domain \cite{li2018pointcnn,atzmon2018point,thomas2019kpconv,liu2019relation} or developing graph neural networks \cite{simonovsky2017dynamic,shen2018mining,wang2019dynamic,xu2020grid,gao2020graphter} to improve point cloud learning.
In this paper, we focus on PointNet \cite{qi2017pointnet}, PointNet++ \cite{qi2017pointnet++} and DGCNN \cite{wang2019dynamic} since these 3D models are extensively involved with practical 3D applications.

\noindent \textbf{Adversarial Attack on 3D Point Clouds.}
Deep neural networks are vulnerable to adversarial examples, which has been extensively explored in the 2D image domain \cite{moosavi2016deepfool,moosavi2017universal}.
Recently, many works \cite{xiang2019generating,wicker2019robustness,zhang2019adversarial,zheng2019pointcloud,tsai2020robust,zhao2020isometry,zhou2020lg,hamdi2020advpc,liu2021imperceptible} adapt adversarial attacks into the 3D vision community, which can be divided into two categories:
1) point adding/dropping attack:
Xiang \textit{et al.} \cite{xiang2019generating} proposed point generation attacks by adding a limited number of synthesized points/clusters/objects to a point cloud.
Recently, more works
\cite{zhang2019adversarial,wicker2019robustness,zheng2019pointcloud} utilize gradient-guided attack methods to identify critical points from point clouds for deletion.
2) point perturbation attack:
Previous point-wise perturbation attacks \cite{wen2020geometry,tsai2020robust} learn to perturb xyz coordinates of each point by adopting the C\&W framework \cite{carlini2017towards} based on the Chamfer and Hausdorff distance with additional consideration of the benign distribution of points.
Later works \cite{hamdi2020advpc,liu2019extending,ma2020efficient,zhang2019defense} further apply the iterative gradient method to achieve more fine-grained adversarial perturbation.

\noindent \textbf{Spectral Methods for Point Clouds}
In the graph spectral domain, the rough shape of a point cloud will be encoded into low-frequency components, which is suitable for denoising point clouds.
Rosman \textit{et al.} proposed spectral point cloud denoising based on the non-local framework \cite{rosman2013patch}. 
They group similar surface patches into a collaborative patch and perform shrinkage in the GFT domain by a low-pass filter, which leads to denoising of the 3D shape. 
Zhang \textit{et al} proposed a tensor-based method to estimate hypergraph spectral components and frequency coefficients of point clouds, which can be used to denoise 3D shapes \cite{zhang2020hypergraph}. 
On the contrary, high-frequency components often represent fine details of point clouds, which can be used to detect contours or process redundant information. 
Chen \textit{et al.} proposed a high-pass filtering-based resampling method to highlight contours for large-scale point cloud visualization and extract key points for accurate 3D registration \cite{chen2017fast}. 
Sameera \textit{et al.} proposed Spectral-GANs to generate high-resolution 3D point clouds, which takes the advantage of spectral representations for compact representation \cite{ramasinghe2020spectral}. 

\section{Graph Spectral Analysis for Point Clouds}
\label{sec:analysis}
In this section, we provide graph spectral analysis for point clouds, which lays the foundation for the proposed graph spectral domain attack in Sec.~\ref{sec:attack}. 
We first discuss the benefit of graph spectral representations of point clouds and introduce how to transform point clouds onto the spectral domain in Sec.~\ref{subsec:transfer}. 
Then, we analyze the roles of different spectral components with respect to geometric structures in Sec.~\ref{subsec:spectral_analysis}.

\subsection{Spectral Representations of Point Clouds}
\label{subsec:transfer}
Signals can be compactly represented in the spectral domain, provided that the transformation basis characterizes principle components of the signals. 
For instance, images are often transformed onto the Discrete Cosine Transform (DCT) domain for compression and processing \cite{choi2020task,li2018learning}.
Different from images supported on regular grids, point clouds reside on irregular domains with no ordering of points, which hinders the deployment of traditional transformations such as the DCT. 
Though we may quantize point clouds onto regular voxel grids or project onto a set of depth images from multiple viewpoints, this would inevitably introduce quantization loss.
Instead, graphs serve as a natural representation for irregular point clouds, which is accurate and structure-adaptive \cite{hu2021overview}. 
With an appropriately constructed graph that well captures the underlying structure, the Graph Fourier Transform (GFT) will lead to a compact representation of geometric data including point clouds in the spectral domain \cite{shen2010edge,hu2012depth,hu2014multiresolution,hu2015intra,zhang2014point,xu2019predictive}, which inspires new insights and understanding of point cloud attacks.   

Formally, we represent a point cloud $\mP=\{\vp_i\}_{i=1}^n \in \mathbb{R}^{n \times 3}$ consisting of $n$ points over a graph $ \gG=\{\gV,\gE, \mA\} $, which is composed of a vertex set $ \gV $ of cardinality $|\gV|=n$ representing points, an edge set $ \gE $ connecting vertices, and an adjacency matrix $\mA$. 
Each entry $a_{i,j}$ in $\mA$ represents the weight of the edge between vertices $i$ and $j$, which often captures the similarity between adjacent vertices. Here, we construct an unweighted $K$-nearest-neighbor graph ($K$-NN graph), where each vertex is connected to its $K$ nearest neighbors in terms of the Euclidean distance with weight $1$. 
The coordinates of points in $\mP$ are treated as graph signals. 

Prior to the introduction of the GFT, we first define the combinatorial graph Laplacian matrix \cite{shuman2013emerging} as $\mL:=\mD-\mA $, where $ \mD $ is the \textit{degree matrix}---a diagonal matrix where $ d_{i,i} = \sum_{j=1}^n a_{i,j} $. Given real and non-negative edge weights in an undirected graph, $\mL$ is real, symmetric, and positive semi-definite \cite{chung1997spectral}. 
Hence, it admits an eigen-decomposition $\mL = \mU \mLambda \mU^{\top}$, where $\mU=[\vu_1,...,\vu_n]$ is an orthonormal matrix containing the eigenvectors $\vu_i$, and $\mLambda = \mathrm{diag}(\lambda_1,...,\lambda_n)$ consists of eigenvalues $\{\lambda_1=0 \leq \lambda_2 \leq ... \leq \lambda_n\}$. 
We refer to the eigenvalues as the {\it graph frequency/spectrum}, with a smaller eigenvalue corresponding to a lower graph frequency.

For any graph signal $\vx \in \mathbb{R}^{n}$ residing on the vertices of $\mathcal G$, its GFT coefficient vector $\hat{\vx}  \in \mathbb{R}^{n}$ is defined as \cite{hammond2011wavelets}:
\begin{equation}
    \hat{\vx}  = \phi_{\text{GFT}}(\vx) = \mU^{\top} \vx. 
\end{equation}
The inverse GFT (IGFT) follows as:
\begin{equation}
    \vx = \phi_{\text{IGFT}}(\hat{\vx}) = \mU \hat{\vx}. 
\end{equation}
Since $\mU$ is an orthonormal matrix, both GFT and IGFT operations are lossless.

\begin{figure*}[t]
\begin{center}
    \includegraphics[width=\textwidth]{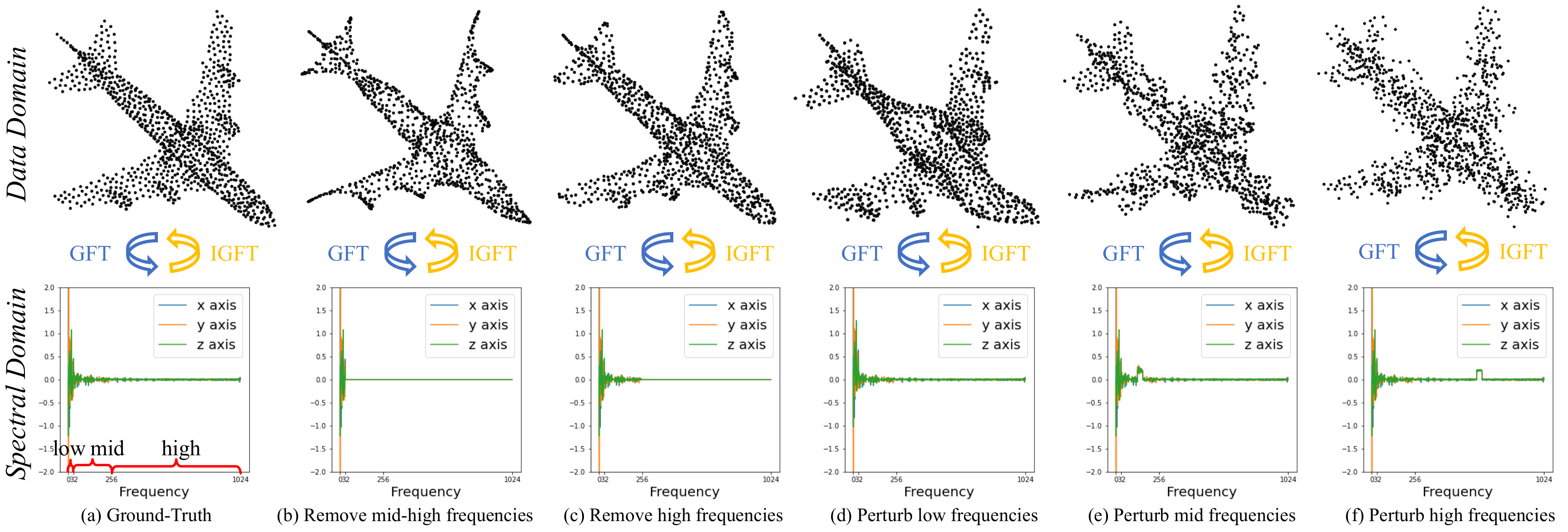}
\end{center}
\vspace{-14pt}
\caption{{ Graph spectral analysis for 3D point clouds.} We take an example from the \textit{airplane} object to investigate the roles of difference frequency bands in the graph spectral domain. (a) Ground-Truth; (b) Remove mid-high frequencies; (c) Remove high frequencies; (d) Perturb low frequencies; (e) Perturb mid frequencies; (f) Perturb high frequencies.}
\label{fig:spectral}
\vspace{-14pt}
\end{figure*}
\vspace{-6pt}
\subsection{Analysis in Graph Spectral Domain}
\label{subsec:spectral_analysis}
\vspace{-6pt}
When an appropriate graph is constructed that captures the geometric structure of point clouds well, the low-frequency components of the corresponding GFT characterize the {\it rough shape} of point clouds, while the high-frequency components represent {\it fine details or noise} (\ie, large variations such as geometric contours) in general.  

To analyze such characteristics, we provide a toy experiment to investigate the roles of difference frequency bands in the graph spectral domain. 
We randomly take a clean point cloud \textit{airplane} from the ModelNet40 dataset \cite{wu20153d}, and sample it into $1024$ points as an example point cloud $\bm{P}$. 
We construct a $K$-NN graph with $K=10$ over the point cloud, and perform the GFT on the three coordinate signals of each point in $\bm{P}$. 
The resulting transform coefficient vectors $\phi_{\text{GFT}}(\bm{P})$ are presented in Figure~\ref{fig:spectral}(a). 
We see that, $\phi_{\text{GFT}}(\bm{P})$ has larger amplitudes at lower-frequency components and much smaller amplitudes at higher-frequency components, demonstrating that most information is concentrated in low-frequency components. 

To further investigate how each frequency band contributes to the geometric structure of the point cloud in the data domain, we first divide the whole spectral domain into three bands: low-frequency band $B_l \in [0,\lambda_l)$, mid-frequency band $B_m \in [\lambda_l,\lambda_h)$, and high-frequency band $B_h \in [\lambda_h,\lambda_{\text{max}}]$, where $\lambda_l,\lambda_h,\lambda_{\text{max}}$ bound the three bands.
While the division of frequency bands is not established, we propose an appropriate division by the distribution of energy---squared sum of transform coefficients. 
In this example, we compute that the lowest 32 frequencies account for almost 90\% of energy, while the lowest 256 frequencies account for almost 97\% of energy.
Based on this observation, we set $\lambda_l=\lambda_{32}=1.19$ and $\lambda_h=\lambda_{256}=13.93$ in this instance.

Next, we study the influence of each frequency band on the geometric structure by removing different bands.
As shown in Figure~\ref{fig:spectral}(b), when GFT coefficients in both mid- and high-frequency bands are assigned $0$, the point cloud reconstructed with only low-frequency components exhibits the rough shape of the original object. 
By adding more information from the mid-frequency band, the reconstructed point cloud has richer local contexts in Figure~\ref{fig:spectral}(c), but still lacks fine-grained details such as the engines of the airplane. 
To summarize, each frequency band is crucial to represent different aspects of the geometric structure of a point cloud. 

Then here is a key question: what are the results of attacking different frequency bands? 
We investigate into this by separately perturbing each frequency band with a large perturbation size, \ie, we perturb consecutive 32 frequencies in each frequency band by adding noise of 0.2 on each frequency.
As shown in Figure~\ref{fig:spectral}(d-f), attacking low-frequency components introduces deformation in the coarse shape but remains smoothness of the surface. 
When attacking mid-frequency components, the shape of the object becomes much rougher. 
In comparison, perturbing high-frequency components loses local details and induces noise and outliers, though the silhouette of the shape is preserved to some degree thanks to the clean lower-frequency components.  

Inspired by the above properties of point clouds in the graph spectral domain, we summarize several insights for developing effective spectral domain attacks:
\begin{itemize}
    \item Each frequency band represents the geometric structure of the point cloud from different perspectives (\eg, low-frequency components represent the basic shape while high-frequency components encode find-grained details). Perturbing only one frequency band may result in the corresponding distortion in the data domain.
    \item Although a large perturbation size can ensure a high success attack rate, it may severely change the spectral characteristics, thus leading to perceptible deformations in the data domain.
\end{itemize}

Based on the above insights, a desirable spectral domain attack for point clouds need to not only perform the perturbation among appropriate frequencies for striking a balance, but also restrict the perturbation size for preserving the original spectral characteristics.




\vspace{-6pt}
\section{Graph Spectral Domain Attack}
\label{sec:attack}
\vspace{-6pt}
\subsection{The Formulation}
\vspace{-6pt}
Given a clean point cloud $\bm{P}=\{\bm{p}_i\}_{i=1}^n \in \mathbb{R}^{n \times 3}$ where each point $\bm{p}_i \in \mathbb{R}^3$ is a vector that contains the (x, y, z) coordinates, a well-trained classifier $f(\cdot)$ can predict its accurate label $y=f(\bm{P})\in \mathbb{Y},\mathbb{Y}=\{1,2,3,...,c\}$ that represents its object class best, where $c$ is the number of classes.
The goal of point cloud attacks on classification is to perturb point cloud $\bm{P}$ into an adversarial one $\bm{P}'$, so that $f(\bm{P}')=y'$ (targeted attack) or $f(\bm{P}')\neq y$ (untargeted attack), where $y' \in \mathbb{Y}$ but $y' \neq y$.


\vspace{-14pt}
\begin{figure*}[h]
    \centering
    \includegraphics[width=\textwidth]{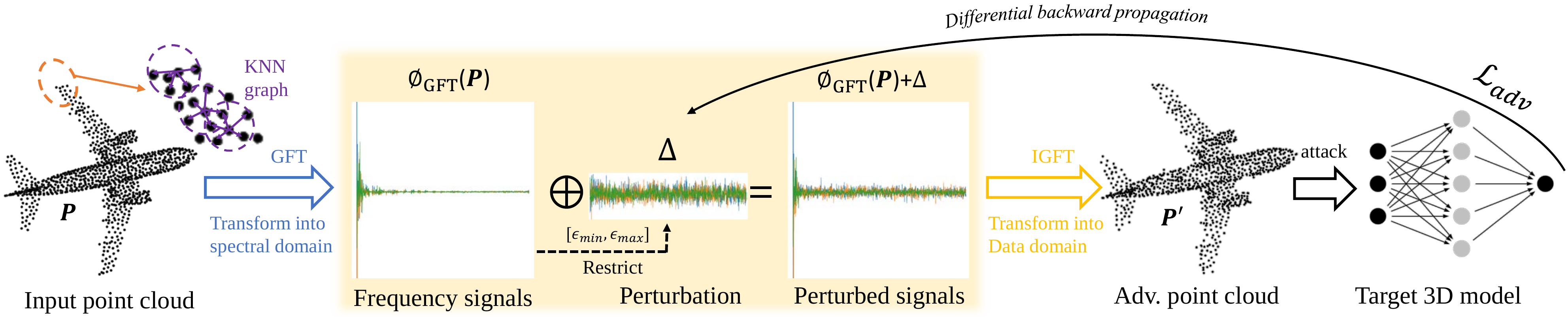}
    \caption{{ The overall pipeline of the proposed GSDA.} Given a clean point cloud, we first construct a $K$-NN graph and transform the point cloud into the spectral domain via the GFT. Then, we perturb the GFT coefficients in a learnable manner with a specifically designed restrictive function over the spectral energy. Subsequently, we transform the perturbed spectral signals back to the data domain via the IGFT. Finally, we take the reconstructed point cloud as the adversarial example and feed it into the target 3D model for attack.}
    \vspace{-14pt}
    \label{fig:pipeline}
\end{figure*}

We propose a novel GSDA attack that aims to learn destructive yet imperceptible perturbations in the spectral domain for generating adversarial point clouds.
Formally, we formulate the proposed GSDA as the following optimization problem:
\begin{equation}
\label{eq:objective}
\begin{aligned}
    & \min_{\bm{\Delta}} \mathcal{L}_{adv}(\bm{P}',\bm{P},y), \text{s.t.} ||\phi_{\text{GFT}}(\bm{P}')-\phi_{\text{GFT}}(\bm{P})||_p < \epsilon, \\
    & \text{where} \ \bm{P}' = \phi_{\text{IGFT}}(\phi_{\text{GFT}}(\bm{P})+\bm{\Delta}),
\end{aligned}
\end{equation}
where $\mathcal{L}_{adv}(\bm{P}',\bm{P},y)$ is the adversarial loss and $\bm{\Delta}$ is the perturbation. 
In the imposed constraint, $\epsilon$ aims to restrict the perturbation size in the spectral domain, which preserves the original spectral characteristics so that the resultant adversarial point cloud $\bm{P}'$ is visually indistinguishable from its clean version $\bm{P}$. 
We adopt the $l_2$-norm in this equation.

To back-propagate the gradient in a desired direction for optimizing the perturbation learning, we define our adversarial loss $\mathcal{L}_{adv}(\bm{P}',\bm{P},y)$ as follows:
\begin{equation}
\label{eq:adv_loss}
    \mathcal{L}_{adv}(\bm{P}',\bm{P},y) = \mathcal{L}_{class}(\bm{P}',y) + \beta \cdot \mathcal{L}_{reg}(\bm{P}',\bm{P}),
\end{equation}
where $\mathcal{L}_{class}(\bm{P}',y)$ is to promote the misclassification of point cloud $\bm{P}'$. $\mathcal{L}_{reg}(\bm{P}',\bm{P})$ is a regularization term that minimizes the distance between $\bm{P}'$ and $\bm{P}$ to guide perturbation in appropriate frequencies. $\beta$ is a penalty parameter controlling the regularization term.

Specifically, $\mathcal{L}_{class}(\bm{P}',y)$ is formulated as a cross-entropy loss as follows:
\begin{equation}
    \mathcal{L}_{class}(\bm{P}',y) = \left\{
    \begin{aligned}
        -\log(p_{y'}(\bm{P}')), \quad &\text{for targeted attack,} \\
        \log(p_{y}(\bm{P}')), \quad &\text{for untargeted attack,}
    \end{aligned}
    \right.
\end{equation}
where $p(\cdot)$ is the softmax functioned on the output of the target model, \ie, the probability with respect to adversarial class $y'$ or clean class $y$. 
By minimizing this loss function, our GSDA optimizes the spectral perturbation $\bm{\Delta}$ to mislead the target model $f(\cdot)$.

Besides, to strike a balance of perturbing different frequency bands, we utilize both Chamfer distance loss \cite{fan2017point} and Hausdorff distance loss \cite{huttenlocher1993comparing} as the $\mathcal{L}_{reg}(\bm{P}',\bm{P})$ function.
This reflects the imperceptibility in the data domain for updating the frequency perturbation $\bm{\Delta}$.
\vspace{-6pt}
\subsection{The Algorithm}
\vspace{-6pt}
Based on the problem formulation, we develop an efficient and effective algorithm for the GSDA model.  
As shown in Figure~\ref{fig:pipeline}, the proposed GSDA attack is composed of four steps:
Firstly, the GSDA transforms the clean point cloud $\bm{P}$ from the data domain to the graph spectral domain via the GFT operation $\phi_{\text{GFT}}$. 
Then, the GSDA perturbs the GFT coefficients through our designed perturbation strategy imposed with an energy constraint function as in Eq.~(\ref{eq:objective}). 
Next, we convert the perturbed spectral signals back to the data domain via the IGFT operation $\phi_{\text{IGFT}}$ for constructing the adversarial point cloud $\bm{P}'$. 
Finally, we optimize the adversarial loss function to iteratively update the desired perturbations added in the spectral domain. 
In the following, we elaborate on each module in order.

\noindent \textbf{Transform onto Spectral Domain.}
Given a clean point cloud $\mP$, we employ the GFT to transform $\mP$ onto the graph spectral domain. 
Specifically, we first construct a $K$-NN graph on the whole point cloud, and then compute the graph Laplacian matrix $\mathbf{L}$. 
Next, we perform eigen-decomposition to acquire the orthonormal eigenvector matrix $\bm{U}$, which serves as the GFT basis. 
The GFT coefficients $\phi_{\text{GFT}}(\bm{P})$ is then obtained by:
\begin{equation}
    \phi_{\text{GFT}}(\bm{P}) = \bm{U}^{\top} \bm{P},
\end{equation}
where $\phi_{\text{GFT}}(\bm{P}) \in \mathbb{R}^{n \times 3}$ corresponds to the transform coefficients of the x-, y-, z-coordinate signals.

\noindent \textbf{Perturbation in the Graph Spectral Domain.}
We deploy a trainable perturbation $\bm{\Delta}$ to perturb the spectral representation of $\bm{P}$:  $\phi_{\text{GFT}}(\bm{P})+\bm{\Delta}$.
Further, to restrict the perturbation size among appropriate frequencies for enhancing the imperceptibility, instead of the constraint of the entire energy in Eq.~(\ref{eq:objective}), we constrain the perturbation $\bm{\Delta}$ on each frequency in a valid range $[\epsilon_{min},\epsilon_{max}]$ in order to keep the trend of spectral characteristics, \ie, low-frequency components with large magnitudes and high-frequency components with small magnitudes.
In particular, we define the perturbation size of each frequency as:
\begin{equation}
    \bm{\Delta} / \phi_{\text{GFT}}(\bm{P}) \in [\epsilon_{min},\epsilon_{max}],
\end{equation}
which maintains a certain ratio to each frequency with $\epsilon_{max}=-\epsilon_{min}$. 
In order to adjust the perturbation $\bm{\Delta}$ adaptively and further improve the success rate of the proposed spectral attack, we formulate the attack process as an optimization problem by leveraging the gradients of the target 3D model $f(\cdot)$ via backward-propagation. 
We update the perturbation $\bm{\Delta}$ with the gradients, and learn the perturbation $\bm{\Delta}$ as:
\begin{equation}
\label{eq:update_delta}
\begin{aligned}
    &\bm{\Delta}' \leftarrow \bm{\Delta} - lr \cdot \partial_{\bm{\Delta}}(\mathcal{L}_{adv}(\bm{P}',\bm{P},y)), \\
    &   \text{s.t.} \quad \bm{\Delta} / \phi_{\text{GFT}}(\bm{P}) \in [\epsilon_{min},\epsilon_{max}],
\end{aligned}
\end{equation}
where $lr$ is the learning rate. 

\noindent \textbf{Inverse Transform onto Data Domain.}
After obtaining the perturbed spectral representations, we apply the IGFT to convert the perturbed signals from the spectral domain back to the data domain as:
\begin{equation}
    \bm{P}' = \phi_{\text{IGFT}}(\phi_{\text{GFT}}(\bm{P})+\bm{\Delta}) = \bm{U} (\phi_{\text{GFT}}(\bm{P})+\bm{\Delta}),
\end{equation}
where $\bm{P}'$ is the crafted adversarial point cloud.

\vspace{-6pt}
\section{Experiments}
\vspace{-2pt}
\subsection{Dataset and 3D Models}
\vspace{-4pt}
\noindent \textbf{Dataset.}
We adopt the point cloud benchmark ModelNet40 \cite{wu20153d} dataset in all the experiments. 
This dataset contains 12,311 CAD models from 40 most common object categories in the world. 
Among them, 9,843 objects are used for training and the other 2,468 for testing. 
As in previous works \cite{qi2017pointnet}, we uniformly sample $n=1,024$ points from the surface of each object, and re-scale them into a unit ball. 
For adversarial point cloud attacks, we follow \cite{xiang2019generating,wen2020geometry} and randomly select 25 instances for each of 10 object categories in the ModelNet40 testing set, which can be well classified by the classifiers of interest.

\noindent \textbf{3D Models.}
We select three commonly used networks in 3D computer vision community as the victim models, \ie, PointNet \cite{qi2017pointnet}, PointNet++ \cite{qi2017pointnet++}, and DGCNN \cite{wang2019dynamic}. 
We train them from scratch, and the test accuracy of each trained model is within 0.1\% of the best reported accuracy in their original papers. We generate the adversarial point clouds on each of them, and further explore their transferability among these three models.
\vspace{-6pt}
\subsection{Implementation Details}
\vspace{-4pt}
\noindent \textbf{Experimental Settings.} For generating the adversarial examples, we update the frequency perturbation $\bm{\Delta}$ with $500$ iterations. We use Adam optimizer \cite{kingma2014adam} to optimize the objective of our proposed GSDA attack in Eq.~(\ref{eq:objective}) with a fixed learning rate $lr=0.01$, and the momentum is set as $0.9$.
We set $K=10$ to build a $K$-NN graph. The penalty $\beta$ in Eq.~(\ref{eq:adv_loss}) is initialized as 10 and adjusted by 10 runs of binary search~\cite{madry2017towards}. The weights of Chamfer distance loss \cite{fan2017point} and Hausdorff distance loss \cite{huttenlocher1993comparing} in the regularization term are set to $5.0$ and $0.5$, respectively.
Since the targeted attack is more challenging than the untargeted attack, we focus on the targeted attack in the experiments.
All experiments are implemented on a single NVIDIA RTX 2080Ti GPU.

\noindent \textbf{Evaluation Metrics.}
To quantitatively evaluate the effectiveness of our proposed GSDA attack, we measure by the attack success rate, which is the ratio of successfully fooling a 3D model. 
Besides, to measure the perturbation size of different attackers, we adopt four evaluation metrics: 
(1) Data domain: $l_2$-norm distance $\mathcal{D}_{norm}$ \cite{cortes2012l2} which measures the square root of the sum of squared shifting distance, 
Chamfer distance $\mathcal{D}_{c}$ \cite{fan2017point} which measures the average squared distance between each adversarial point and its nearest original point, 
Hausdorff distance $\mathcal{D}_{h}$ \cite{huttenlocher1993comparing} which measures the maximum squared distance between each adversarial point and its nearest original point and is thus sensitive to outliers; 
(2) Spectral domain: perturbed energy $\mathcal{E}_{\Delta}=||\phi_{\text{GFT}}(\bm{P}')-\phi_{\text{GFT}}(\bm{P})||_2$. 

\begin{figure*}[t]
\begin{center}
    \includegraphics[width=\textwidth]{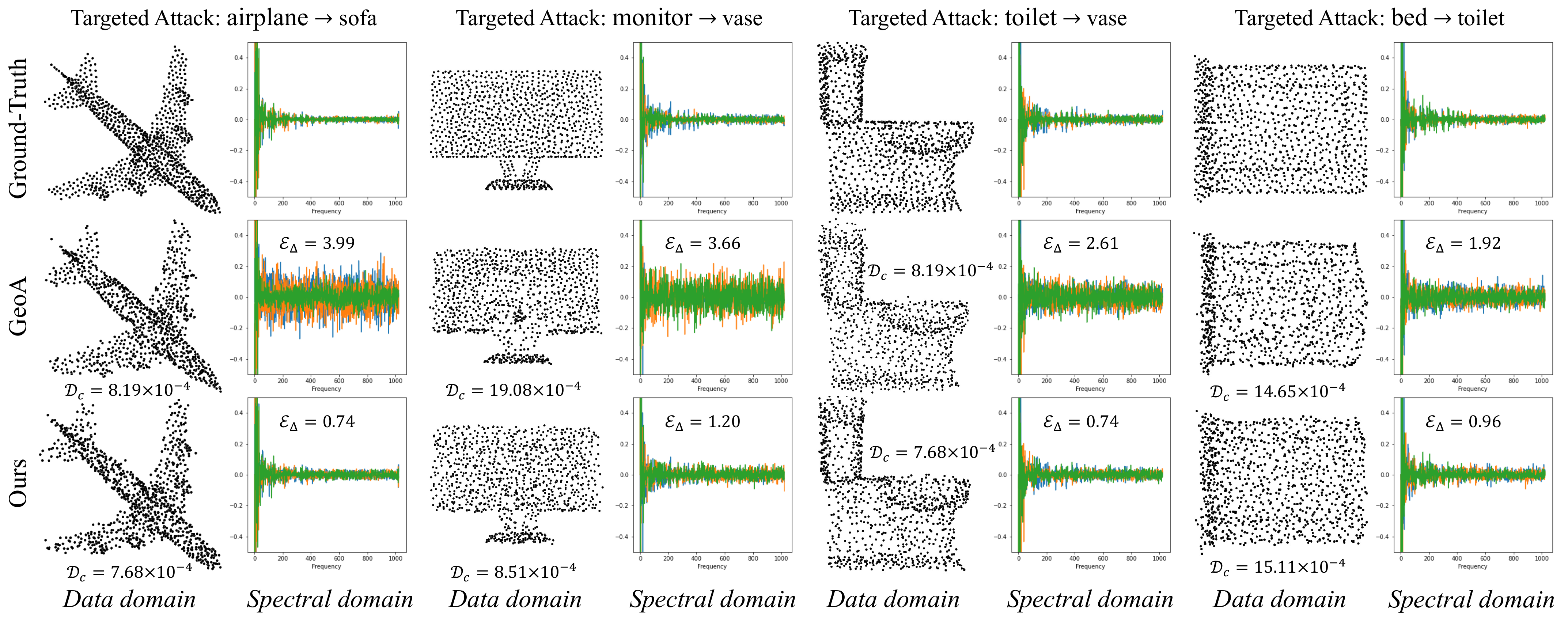}
\end{center}
\vspace{-14pt}
\caption{Visualization of the generated adversarial examples in both the data and spectral domains. Specifically, we compare our GSDA with GeoA via evaluation metrics of the perturbation budget $\mathcal{D}_c$ in the data domain and the perturbed energy $\mathcal{E}_{\Delta}$ in the spectral domain. }
\label{fig:visualize_contrast}
\vspace{-14pt}
\end{figure*}

\subsection{Evaluation on Our GSDA Attack}
\begin{table}[]
\vspace{-20pt}
\centering
\caption{Comparative results on the perturbation sizes of different methods in the data domain for adversarial point clouds. The results of 3D-ADV are borrowed from its original paper. $^p, ^c, ^o$ denote the variant of adversarial point, adversarial cluster and adversarial object, respectively.
}
\vspace{-10pt}
\scalebox{0.75}{
\begin{tabular}{cccccc}
\hline
\multirow{2}{*}{\begin{tabular}[c]{@{}c@{}}Attack\\ Model\end{tabular}} & \multirow{2}{*}{Methods} & \multirow{2}{*}{\begin{tabular}[c]{@{}c@{}}Success\\ Rate\end{tabular}} & \multicolumn{3}{c}{Perturbation Size} \\ \cline{4-6} 
                                                                        &                          &                                                                         & $\mathcal{D}_{norm}$     & $\mathcal{D}_{c}$       & $\mathcal{D}_{h}$       \\ \hline
\multirow{4}{*}{PointNet}                                               & FGSM                     & 100\%                                                                   & 0.7936      & 0.1326     & 0.1853     \\
                                                                        & 3D-ADV$^p$                   & 100\%                                                                   & 0.3032      & \textbf{0.0003}     & 0.0105     \\
                                                                         & 3D-ADV$^c$ & 92.1\% & - & 0.1652 & - \\
     & 3D-ADV$^o$ & 81.9\% & - & 0.1321 & - \\
                                                                        & GeoA                     & 100\%                                                                   & 0.4385      & 0.0064     & 0.0175     \\
                                                                        & Ours                     & 100\%                                                                   &     \textbf{0.1741}        & 0.0007              & \textbf{0.0031}     \\ \hline
\multirow{4}{*}{PointNet++}                                             & FGSM                     & 100\%                                                                   & 0.8357      & 0.1682     & 0.2275     \\
                                                                        & 3D-ADV$^p$                    & 100\%                                                                   & 0.3248      & \textbf{0.0005}     & 0.0381     \\
                                                                        & GeoA                     & 100\%                                                                   & 0.4772      & 0.0198     & 0.0357     \\
                                                                        & Ours                     & 100\%                                                                   &     \textbf{0.2072}        &     0.0081       & \textbf{0.0248}           \\ \hline
\multirow{4}{*}{DGCNN}                                                  & FGSM                     & 100\%                                                                   & 0.8549      & 0.189      & 0.2506     \\
                                                                        & 3D-ADV$^p$                    & 100\%                                                                   & 0.3326      & \textbf{0.0005}     & 0.0475     \\
                                                                        & GeoA                     & 100\%                                                                   & 0.4933      & 0.0176     & \textbf{0.0402}     \\
                                                                        & Ours                     & 100\%                                                                   &     \textbf{0.2160}        & 0.0104     & 0.1401     \\ \hline
\end{tabular}}
\label{tab:perturbation}
\vspace{-20pt}
\end{table}

\noindent \textbf{Quantitative results.}
In order to fairly compare our GSDA attack with existing methods, we perform four adversarial attacks, namely FGSM \cite{zhang2019adversarial}, 3D-ADV \cite{xiang2019generating}, GeoA \cite{wen2020geometry} and ours, and measure the perturbation in the data domain with three evaluation metrics when these methods reach 100\% of attack success rate. Specifically, we implement these attacks on three 3D models PointNet, PointNet++ and DGCNN. Corresponding results are shown in Table~\ref{tab:perturbation}.
We see that, our GSDA generates adversarial point clouds with almost the lowest perturbation sizes in all evaluation metrics on three attack models. 

Note that, as $\mathcal{D}_c$ measures the average squared distance between each adversarial point and its nearest original point, attacking by adding a few points in 3D-ADV$^p$ has a natural advantage in terms of $\mathcal{D}_c$ because most of the distance is equal to $0$. However, it induces much larger distortions than ours on the other two metrics.

Overall, this demonstrates that our generated point clouds are less distorted quantitatively.  
Besides, for each attack method, it takes larger perturbation sizes to successfully attack PointNet++ and DGCNN than to attack PointNet, which indicates that PointNet++ and DGCNN are harder to attack.


\noindent \textbf{Visualization results.}
We also provide some visualization results of generated adversarial point clouds and corresponding spectral coefficients for comparison. As shown in Figure~\ref{fig:visualize_contrast}, we compare the visualization of both GeoA and ours on four examples under the setting of targeted attacks. Since GeoA implements perturbation in the data domain while we conduct it in the spectral domain, we provide the perturbation budgets in both two domains for fair comparison.
Figure~\ref{fig:visualize_contrast} shows that our GSDA attack has less perturbation $\mathcal{E}_{\Delta}$ than GeoA in the spectral domain as we develop an energy constraint function for optimization. By reflecting the perturbation in the data domain, our adversarial examples are more imperceptible than those of GeoA in both local details and distributions. Quantitatively, we achieve smaller perturbation budget $\mathcal{D}_c$ in the data domain. To conclude, our GSDA attack is more effective and imperceptible.
\vspace{-12pt}
\subsection{Analysis on Robustness of Our GSDA Attack}
\vspace{-6pt}
\noindent \textbf{Attacking the Defenses.}
To further examine the robustness of our proposed GSDA attack, we employ several 3D defenses to investigate whether our attack is still effective.
Specifically, we employ the PointNet model with the following defense methods: Statistical Outlier Removal (SOR) \cite{zhou2019dup}, Simple Random Sampling (SRS) \cite{zhang2019adversarial}, DUP-Net defense \cite{zhou2019dup} and IF-Defense \cite{wu2020if}. 
Table~\ref{tab:defense_drop} shows that across a range of dropping ratios via the defense SOR, 
the attack success rates drop a lot.
However, the performance of our GSDA attack decays much slower than that of GeoA, validating that our attack is much more robust.
We also report the results with other defenses in Table~\ref{tab:various_defense}.
We observe that FGSM and 3D-ADV attacks have low success rates under all the defenses, which is because they often lead to uneven local distribution and outliers. Besides, GeoA achieves relatively higher attack success rates, since it utilizes a geometry-aware loss function to constrain the similarity in curvature and thus has fewer outliers. In comparison, our attack achieves the highest success rates than all other attacks under all defenses since the trivial perturbation in the spectral domain reflects less noise in the data domain, thus enhancing the robustness.

\noindent \textbf{Transferability of Adversarial Point Clouds.}
\begin{table}[t]
\parbox{.48\linewidth}{
\centering
\vspace{-10pt}
\centering
\caption{Defense by dropping different ratios of points via SOR.}
\vspace{-10pt}
\scalebox{0.65}{
\begin{tabular}{cccccccc}
\hline
\multirow{3}{*}{Method} & \multicolumn{7}{c}{Attack success rate (\%)}                               \\
                        & \multicolumn{7}{c}{defense via SOR} \\ \cline{2-8} 
                        & 0\%     & 1\%       & 2\%      & 5\%      & 10\%     & 15\%     & 20\%     \\ \hline
GeoA                   & \textbf{100}     & 83.47     & 70.56    & 52.61    & 31.58    & 18.62    & 11.71    \\
Ours                    & \textbf{100}     & \textbf{91.87}     & \textbf{89.91}    & \textbf{85.38}    & \textbf{72.53}    & \textbf{50.00}    & \textbf{27.78}    \\ \hline
\end{tabular}
}
\label{tab:defense_drop}
}
\hfill
\parbox{.48\linewidth}{
\centering
\caption{The attack success rate (\%) on PointNet model by various attacks under defense.}
\vspace{-10pt}
\scalebox{0.75}{
\begin{tabular}{ccccc}
\hline
Attack & No Defense & SRS   & DUP-Net & IF-Defense \\ \hline
FGSM   & 100\%        & 9.68\%  & 4.38\%    & 4.80\%        \\
3D-ADV & 100\%        & 22.53\% & 15.44\%   & 13.70\%       \\
GeoA   & 100\%        & 67.61\% & 59.15\%   & 38.72\%      \\
Ours   & 100\%        & \textbf{81.03\%}    &  \textbf{68.98\%}       &   \textbf{50.26\%}         \\ \hline
\end{tabular}}
\label{tab:various_defense}
}
\vspace{-20pt}
\end{table}
To investigate the transferability of our proposed GSDA attack, we craft adversarial point clouds on normally trained models and test them on all the three 3D models we consider. The success rates, which are the misclassification rates of the corresponding models on adversarial examples, are shown in Table~\ref{tab:transfer}. 
The left, middle and right parts present the three models we attack: PointNet, PointNet++ and DGCNN; the columns of the table present the models we test.
We observe that although our GSDA attack is not tailored for transferability, it has relatively higher success rates of transfer-based attacks than others. This is because our perturbation in the spectral domain not only keeps spectral characteristics but also reflects trivial noise in the data domain, which is thus more robust. A further improvement of the transferability could be developed by employing an additional adversarial learning mechanism with the Auto-Encoder reconstruction as in \cite{hamdi2020advpc} (left as the future works).
The comparatively slightly lower transferability of our GSDA in this table may be related with special properties of certain 3D models. 
To summarize, this intrinsic property makes it possible to design black-box defense against such adversarial instances.

\vspace{-10pt}
\subsection{Ablation Study}
\begin{table}[t]
\vspace{-10pt}
\centering
\caption{The attack success rate (\%) of transfer-based attacks.}
\vspace{-10pt}
\scalebox{0.7}{
\begin{tabular}{c|ccc|ccc|ccc}
\hline
                              Attacks & PointNet & PointNet++ & DGCNN & PointNet & PointNet++ & DGCNN & PointNet & PointNet++ & DGCNN \\ \hline
                              FGSM    & \textbf{100\%}      & 3.99\%                              & 0.63\%     & 3.16\%     & \textbf{100\%}                               & 5.57\%          & 3.59\%     & 7.21\%                              & \textbf{100\%}                          \\
                              3D-ADV  & \textbf{100\%}      & 8.45\%                              & 1.28\%     & 6.63\%     & \textbf{100\%}                               & 10.98\%           & 6.82\%     & 13.53\%                             & \textbf{100\%}                          \\
                              GeoA    & \textbf{100\%}      & \textbf{11.59\%}                             & 2.59\%     & 9.47\%     & \textbf{100\%}                               & 19.77\%           & 12.46\%    & 24.24\%                             & \textbf{100\%}                          \\
 Ours  & \textbf{100\%}      & 11.51\%                                  &   \textbf{8.39\%}                           & \textbf{10.89\%}         & \textbf{100\%}                               &           \textbf{30.84\%}      &     \textbf{32.31\%}     &        \textbf{84.49\%}                           & \textbf{100\%}                          \\ \hline  
\end{tabular}}
\label{tab:transfer}
\vspace{-20pt}
\end{table}
\begin{table}[]
\centering
\vspace{-20pt}
\caption{Sensitivity analysis on the number $K$.}
\vspace{-10pt}
\scalebox{0.80}{
\begin{tabular}{ccccc}
\hline
Number $K$ & Success Rate & $\mathcal{D}_c$ & $\mathcal{D}_h$ & $\mathcal{E}_{\Delta}$ \\ \hline
$K$=5 & 100\% & \textbf{0.0007}  & 0.0031 & 3.4679 \\
$K$=10 & 100\% & \textbf{0.0007} & 0.0031 & \textbf{3.4510} \\
$K$=20 & 100\% & \textbf{0.0007} & \textbf{0.0030} & 3.4617 \\
$K$=40 & 100\% & \textbf{0.0007} & \textbf{0.0030} & 3.4688 \\ \hline
\end{tabular}}
\label{tab:ablation_K}
\vspace{-20pt}
\end{table}
\begin{table}[]
\vspace{-20pt}
\centering
\caption{Comparison of different spectral representations.}
\vspace{-10pt}
\scalebox{0.85}{
\begin{tabular}{cccccccc}
\hline
\multirow{3}{*}{Method} & \multicolumn{4}{c}{Attack success rate (\%)} & \multirow{3}{*}{$\mathcal{D}_{c}$} & \multirow{3}{*}{$\mathcal{D}_{h}$} & \multirow{3}{*}{$\mathcal{E}_{\Delta}$}   \\
                        & \multicolumn{4}{c}{defense via SOR}          &                     &                     &                      \\ \cline{2-5}
                        & 0\%      & 5\%       & 10\%      & 20\%      &                     &                     &                      \\ \hline
Ours-3DDCT                & 93.06      &  37.48    &   25.17   &    19.60    &   0.0029            &   0.0415            &  19.4804  \\
Ours-GFT                    & \textbf{100.00}      & \textbf{85.38}     & \textbf{72.53}     & \textbf{27.78}     & \textbf{0.0007}              & \textbf{0.0031}              & \textbf{3.4510}  \\ \hline
\end{tabular}}
\label{tab:dct}
\vspace{-15pt}
\end{table}
\noindent \textbf{Choice of Spectral Representation.}
We first evaluate the benefit of spectral representation in the GFT domain.  
Since the DCT is widely used in the 2D field to transform images onto the spectral domain, we compare the spectral domain attack performances on different spectral representations of the DCT and GFT. 
We implement 1D DCT and 3D DCT for comparison. 
For 1D DCT, we regard coordinates as the signal and perform 1D DCT on it.
As to 3D DCT, we voxelize the pointcloud and perform 3D DCT on coordinates signal.
As shown in Table~\ref{tab:dct}, we see that our attack with the GFT is more robust to SOR defense and has much lower perturbation size in the spectral domain than the attack with the DCT.
The main reason is that point clouds are unordered, which makes it challenging to capture the correlations among points by the DCT. 
In contrast, the GFT well captures the underlying structure of point clouds via the appropriate graph construction.

\noindent \textbf{Sensitivity on the Number $K$.}
As shown in Table~\ref{tab:ablation_K}, we investigate whether the adversarial effects vary with respect to different settings of the number $K$ in the $K$-NN graph.
Specifically, we implement $K=5,10,20,40$ to perform the proposed GSDA attack on PointNet model and report the corresponding perturbation budgets when achieving 100\% of attack success rate. We see that the attack performance is insensitive to the number $K$ since our attack with different $K$'s requires similar perturbation budgets $\mathcal{D}_c,\mathcal{D}_h$ in the data domain, as well as in the spectral domain measured by $\mathcal{E}_{\Delta}$.

\begin{figure}[htbp]
\vspace{-10pt}
\centering
\centering
\includegraphics[width=\columnwidth]{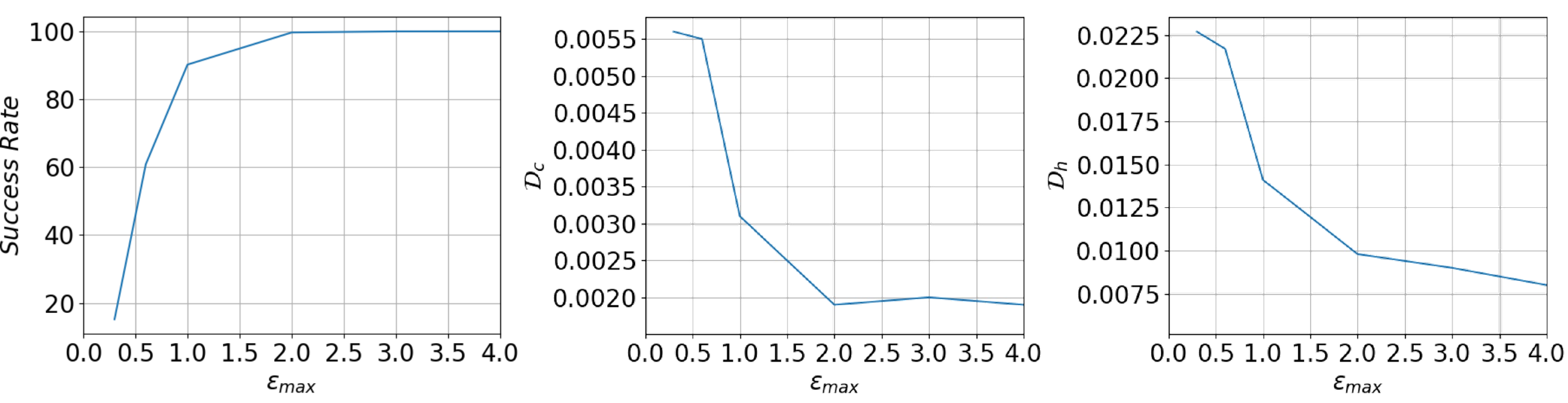}
\vspace{-20pt}
\caption{Attacking performance when applying different $\epsilon_{max}$ restrict, in terms of success rate, Chamfer distance and Hausdorff distance results between adversarial point clouds and the originals.}
\label{fig:epsilon}
\vspace{-30pt}
\end{figure}
\begin{figure}[htbp]
\centering
\includegraphics[width=0.7\columnwidth]{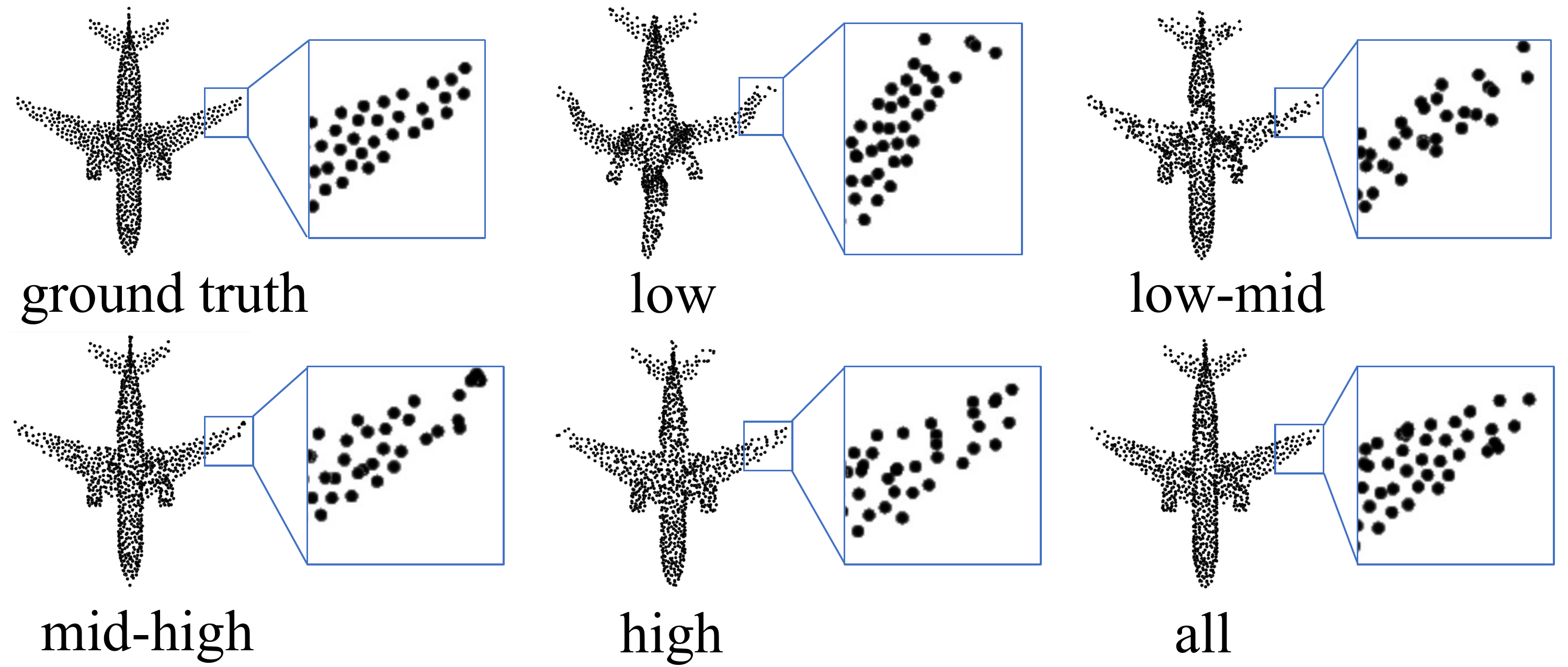}
\caption{Visualization of attacks on specific frequency bands.}
\label{fig:frequency_bands}
\vspace{-20pt}
\end{figure}
\noindent \textbf{Sensitivity on the Perturbation Size $[\epsilon_{min},\epsilon_{max}]$.} As shown in Figure~\ref{fig:epsilon}, when we relax the perturbation constraint $[\epsilon_{min},\epsilon_{max}]$ in the spectral domain, the success rates of our GSDA increase and achieve almost 100\% of success rate when $\epsilon_{max}=-\epsilon_{min}=3.0$, meanwhile $\mathcal{D}_c$ and $\mathcal{D}_h$ achieve an good performance.  

\noindent \textbf{Attacks on Specific Frequency Bands.}
We also perform an ablation study on how attacking specific frequency bands (\ie, low, low-mid, mid-high, high) affects point clouds in the data domain.
As shown in Figure~\ref{fig:frequency_bands}, only attacking low or high frequency band results in distortion of the rough shape or local details, which is consistent with our analysis in Sec.~\ref{sec:analysis}. Instead, attacking appropriate positions among the entire spectra achieves the most imperceptible results.
\vspace{-12pt}
\section{Conclusion}
\vspace{-6pt}
We propose a novel paradigm of point cloud attacks---Graph Spectral Domain Attack (GSDA), which explores insightful spectral characteristics and performs perturbation in the spectral domain to generate adversarial examples with geometric structures well preserved. 
Extensive experiments show the vulnerability of popular 3D models to the proposed GSDA and demonstrate the robustness of our adversarial point clouds.
Such spectral methodology blazes a new path of developing attacks in 3D deep learning.  
Future promising directions include developing effective defense methods against such spectral domain attacks.


\clearpage
%
%
\bibliographystyle{splncs04}
\bibliography{egbib}
\end{document}